%% file: main.tex
\documentclass{article}

\usepackage{microtype}
\usepackage{graphicx}
\usepackage{booktabs}
\usepackage{colortbl}
\usepackage{amsmath}
\usepackage{tikz}
\usetikzlibrary{arrows.meta}

\usepackage[accepted]{hibana-mlsys2025}
\usepackage[hyperfootnotes=false]{hyperref}
\usepackage{xurl}
\definecolor{mydarkblue}{rgb}{0,0.08,0.45}
\makeatletter
\renewcommand{\Notice@String}{Preprint.}
\makeatother
\hypersetup{
  colorlinks=true,
  allcolors=mydarkblue,
  pdfborder={0 0 0},
  pdfpagemode=UseNone,
  pdfview=FitH,
  pdfsubject={MLSys 2027},
  pdfauthor={Yahya Emara, Chi-Chih Chang, Anthony Fei, Ahmed F AbouElhamayed, Chong-Yan Chen, Hung-Yueh Chiang, Kai-Chiang Wu, Mohamed S. Abdelfattah}}
\input{macros}

\newcommand{\titl}[0]{Hibana: Fast Agentic Inference on Unified-Memory Devices}

\mlsystitlerunning{\titl}
\begin{document}
\twocolumn[
\mlsystitle{\titl}

\mlsyssetsymbol{equal}{*}
\begin{mlsysauthorlist}
\mlsysauthor{Yahya Emara}{equal,cornell}
\mlsysauthor{Chi-Chih Chang}{equal,cornell}
\mlsysauthor{Anthony Fei}{cornell}
\mlsysauthor{Ahmed F AbouElhamayed}{cornell}
\mlsysauthor{Chong-Yan Chen}{uiuc}
\mlsysauthor{Hung-Yueh Chiang}{ind}
\mlsysauthor{Kai-Chiang Wu}{nycu}
\mlsysauthor{Mohamed S. Abdelfattah}{cornell}
\end{mlsysauthorlist}
\mlsysaffiliation{cornell}{Cornell University}
\mlsysaffiliation{uiuc}{Department of Computer Science, University of Illinois Urbana-Champaign}
\mlsysaffiliation{ind}{Independent
Researcher}
\mlsysaffiliation{nycu}{National Yang Ming Chiao Tung University}
\mlsyscorrespondingauthor{Yahya Emara}{ye52@cornell.edu}
\mlsyscorrespondingauthor{Chi-Chih Chang}{cc2869@cornell.edu}
\mlsyskeywords{LLM inference, speculative decoding, quantization, prefix caching}
\vskip 0.1in
{\centering\url{https://sharar-project.github.io}\par}
\vskip 0.25in

\begin{abstract}
\input{sections/abstract}
\end{abstract}
]
\printAffiliationsAndNotice{\mlsysEqualContribution}

\input{sections/introduction_cc}
\input{sections/background}
\input{sections/methodology}
\input{sections/evaluation}
\input{sections/related_work}
\input{sections/conclusion}

\input{main.bbl}
\input{sections/AI_usage_statement}

\clearpage
\appendix
\label{appendix-start}
\input{sections/appendix}
\end{document}

%% file: macros.tex
\usepackage{tcolorbox}
\tcbuselibrary{skins}
\definecolor{takeawaycolor}{HTML}{3E6E6B}
\newtcolorbox{takeaway}[1][Takeaway]{enhanced, colback=takeawaycolor!5,
  colframe=takeawaycolor, boxrule=0.6pt, arc=2mm, title={#1},
  fonttitle=\bfseries\small, coltitle=white,
  attach boxed title to top left={xshift=3mm, yshift*=-\tcboxedtitleheight/2},
  boxed title style={colback=takeawaycolor, colframe=takeawaycolor, arc=1.5mm,
    boxrule=0pt, top=0.5mm, bottom=0.5mm},
  top=2.5mm, bottom=1.5mm, left=2mm, right=2mm, fontupper=\small}

%% file: sections/abstract.tex
Local AI promises capable personal assistants that run on users' own
devices, keeping data and computation under their control.
Open-weight models that fit on local devices can now handle practical tasks such as coding. Yet current inference engines fall short. Popular engines such as vLLM and SGLang target datacenter GPUs with dedicated host memory, while local engines underuse the memory hierarchy of unified-memory devices.
We present Hibana, an inference engine for unified-memory devices that co-optimizes model representation, memory management, and decoding strategy. 
\textbf{Model surgery}, the process of editing a trained model's architecture to suit its hardware, changes how the model computes, shrinking its memory footprint and data movement.
\textbf{Unified-memory management} preserves each agent's state across turns, so long-running agents can resume instantly instead of recomputing their state.
\textbf{Adaptive speculation} generates several tokens per step.
Hibana codesigns these features around the shared memory capacity and limited bandwidth of unified-memory systems-on-chip to support local multi-agent workloads.
On a single DGX Spark serving Qwen and Nemotron models of up to 125B parameters, Hibana completes up to
3.4$\times$ more agentic tasks per hour and delivers up to 3.9$\times$ higher decoding throughput than other established frameworks such as vLLM, NVIDIA Flare, llama.cpp,
and TensorFold. Code is available at
\url{https://github.com/sharar-project/hibana}.

%% file: sections/introduction_cc.tex
\section{Introduction}
\label{sec:introduction}

\begin{figure}[t]
    \centering
    \includegraphics[width=1.0\linewidth]{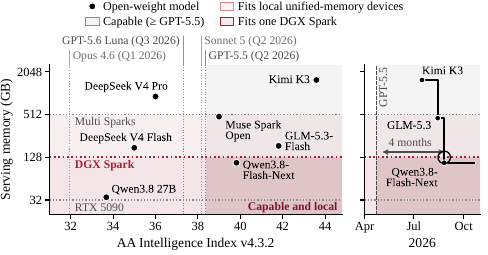}
    \caption{Open-weight models' capability and serving memory. {\textbf{Left}}
    Qwen3.8-Flash-Next scores above GPT-5.5 and fits on one DGX Spark.
    (\textbf{Right.}) The smallest model at GPT-5.5's level \textit{fits} on one
    DGX Spark $\sim$4 months after GPT-5.5 was released.
    }
    \label{fig:trends}
\end{figure}

Large language models have evolved from conversational assistants into systems capable of writing code and executing multi-step agentic tasks. Driven by scaling laws~\cite{kaplan2020scaling}, frontier models have grown to trillions of parameters, making datacenter-scale infrastructure the dominant way to serve them. Running capable models locally would keep user data under their control and eliminate metered API inference charges. Recent progress in both model and hardware suggests that this is becoming practical. Figure~\ref{fig:trends} shows that Qwen3.8-Flash-Next surpasses GPT-5.5 on the Artificial Analysis Intelligence Index~\cite{artificialanalysis2026index} while fitting within the 128 GB memory of a single NVIDIA DGX Spark. Moreover, an open-weight model matching GPT-5.5's capability reached this memory budget approximately four months after GPT-5.5's release.

This progress is driven by improvements in model architecture and training efficiency. Efficient architectures, including mixture-of-experts and sparse or linear attention, reduce inference costs~\cite{fedus2022switch,child2019generating,yang2025gated,dao2024transformers}, while knowledge distillation~\cite{hinton2015distilling,gu2024minillm} and low-precision formats such as \texttt{NVFP4} and \texttt{MXFP4} further reduce serving requirements. Combined with increasingly capable local hardware, these advances bring LLM-powered agents within reach of local devices. However, fitting a capable model into memory does not guarantee efficient execution of agentic workloads, because decoding is limited by memory bandwidth, which unified-memory devices trade for capacity.

\begin{figure*}
    \centering
    \includegraphics[width=1.0\linewidth]{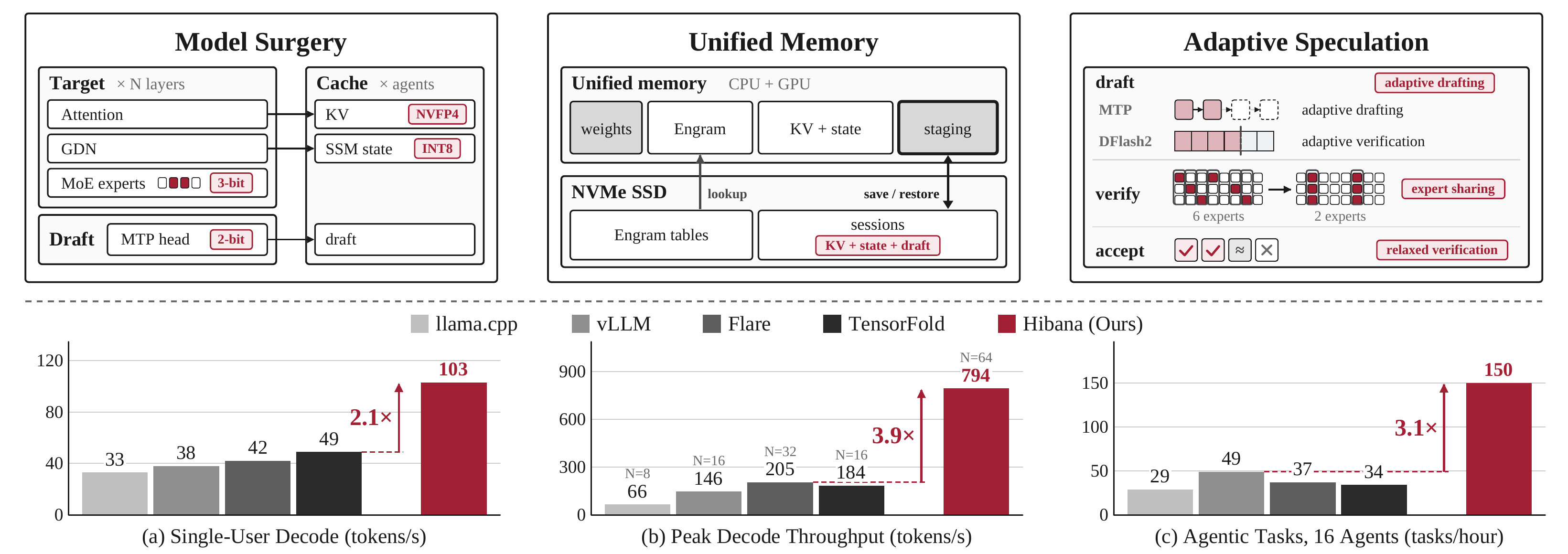}
    \caption{Hibana serving Qwen3.8-Flash-Next on one DGX Spark. (Top.)
    Hibana's model surgery, unified-memory management, and adaptive
    speculation (Section~\ref{sec:design}). (Bottom.) (a) decode speed for a
    single user, (b) peak decode throughput at each engine's best batch
    size, and (c) agentic tasks completed per hour with 16 agents. Speedups
    are over the strongest baseline.}
    \label{fig:teaser}
\end{figure*}

Traditional local inference has largely focused on single-user chat, often with one active decoding stream. Modern agents, however, invoke tools and spawn subagents, generating concurrent requests while maintaining persistent sessions across turns. This concurrency creates opportunities for batching and amortizing model-weight reads, which is particularly important on unified-memory devices with limited memory bandwidth. Meanwhile, persistent sessions compete with active requests for the same memory, and unified-memory devices have no separate host-DRAM tier to move them to.

Existing inference systems address these requirements only partially. llama.cpp~\cite{llamacpp2026} supports extensive sub-4-bit quantization but exhibits limited throughput scaling with increasing batch size in our measurements. vLLM and SGLang~\cite{kwon2023pagedattention,zheng2024sglang} primarily target datacenter GPUs, while recent local systems such as NVIDIA Flare~\cite{nvidia2026flare} and TensorFold~\cite{tensorfold2026} support agentic workloads but leave opportunities to improve expert-weight reuse and session management.

We present \textbf{Hibana}, an inference engine designed for concurrent agentic workloads on unified-memory devices. Hibana jointly optimizes model representation, memory management, and decoding to reduce memory traffic per accepted token. The contributions of Hibana are threefold:

\begin{itemize}
\setlength{\itemsep}{0pt}
\setlength{\parsep}{0pt}
\setlength{\parskip}{0pt}

\item \textbf{Model surgery.} We define \emph{model surgery} as editing a trained model's architecture to suit its hardware. Specifically, we introduce expert sharing, which selects a shared set of experts per layer for batched decoding and verification, and a two-bit draft head with FP8 rescoring, which cuts each draft step's head read from about 640\,MB to about 180\,MB. We also reduce model-weight traffic by encoding routed experts and dense-layer decode copies as three-bit FP8-codebook indices for FP8 tensor-core execution.

\item \textbf{Unified memory management.} We design an \emph{SSD session cache} that preserves agents' KV, recurrent, and draft states across turns using a small, fixed staging buffer. INT8 recurrent states shrink every session Hibana keeps, and \emph{Engram offloading} serves large embedding tables from the SSD.

\item \textbf{Adaptive speculation.} We adapt MTP draft lengths using recent acceptance rates, subject to batch-level execution constraints (adaptive drafting), and adapt the number of diffusion-drafted tokens verified on dense models (adaptive verification). We also optionally accept draft tokens that the target ranks near its top prediction (relaxed verification), bounding the amplification of next-token probabilities.

\end{itemize}

We evaluate Hibana on a single NVIDIA DGX Spark using Qwen3.8-27B~\cite{qwen2026qwen38_27b}, Qwen3.8-Flash-Next~\cite{qwen2026qwen38flashnext}, Nemotron-3.5-Bolt~\cite{nvidia2026nemotron35bolt}, and Nemotron-3.5-Super~\cite{nvidia2026nemotron35super}. We measure agentic tasks completed per hour alongside decoding throughput and time to first token. Compared with vLLM, Flare, llama.cpp, and TensorFold, Hibana completes up to $3.4\times$ more agentic tasks per hour and achieves up to $3.9\times$ higher decoding throughput (Figure~\ref{fig:teaser}).

%% file: sections/background.tex
\section{Background and Challenges}
\label{sec:background}

\subsection{High Capacity, Limited Bandwidth on DGX Spark} 

NVIDIA DGX Spark has 128\,GB of unified LPDDR5X memory shared by the
CPU and GPU. This allows large models to fit in a compact system, but
its 256-bit memory interface provides only 273\,GB/s of peak bandwidth,
far less than the HBM used in datacenter GPUs~\cite{nvidia2026sparkhardware}.
This trade-off matters during autoregressive decoding, which generates
one token at a time. At small batch sizes, each step must read the
weights needed for the next token, so decoding is often limited by
memory bandwidth rather than compute.

Two common approaches to this problem are weight quantization and
speculative decoding. Both help, but each faces challenges on this hardware.

\textbf{Challenge 1. Sub-4-bit weights.} Quantization reduces memory traffic by storing
weights with fewer bits. Going below 4 bits cuts traffic further but makes model quality harder to preserve, and GPUs lack native 2- or 3-bit matrix operations, so weights must be converted to a supported format before each multiply. Without efficient kernels, this conversion can offset the bandwidth savings. Hibana addresses this with calibrated three-bit codebook weights that tensor cores consume as FP8 (\S\ref{sec:weights}).

Speculative decoding takes a different approach: it tries to produce
more tokens from each load of the target model's weights. A cheaper
draft model proposes several tokens, which the target model verifies
together~\cite{leviathan2023speculative}. When multiple tokens are
accepted, the cost of reading the target weights is spread across
those tokens. Its speedup depends on how many draft tokens are accepted and what drafting and verification cost, and on this hardware both work against it.

\textbf{Challenge 2. Expert reads during verification.} On mixture-of-experts (MoE) models each verified token routes to its own experts~\cite{jiang2024mixtral}, so verifying more tokens reads more expert weights. On Qwen3.8-Flash-Next, each token's top five experts alone cover 250 to 460 of a layer's 512 at 16 to 64 requests, so little of the weight reuse that speculation relies on remains. Hibana addresses this with expert sharing (\S\ref{sec:moe}).

\textbf{Challenge 3. Drafting cost and varying acceptance.} Acceptance varies across prompts and requests, so a fixed draft length either wastes drafting and verification or leaves acceptable tokens undrafted. Drafting is also not free on a bandwidth-bound device, because even a small draft module reads a full-vocabulary output head at every step. Hibana addresses this with adaptive speculation (\S\ref{sec:speculation}) and a two-bit draft head (\S\ref{sec:draftcomp}).

\subsection{State Reuse in Agentic Workloads}

Agentic workloads create an additional memory challenge. A coding agent
may pause while waiting for tool calls, while subagents explore
different branches of a shared conversation. These sessions often
share an initial prefix but need their own state to resume. Keeping
every session in memory limits concurrency, while discarding cached
state forces the engine to recompute context when a session returns.

In hybrid attention/recurrent models, resuming a session requires
both its attention key-value (KV) cache and a snapshot of each
recurrent layer's state at the same token position
~\cite{yang2025gated}. Preserving the draft model's state can
also avoid rebuilding its context. Together, these form the cached
state needed to resume an agent's \textit{session}.

\textbf{Challenge 4. Session state in shared memory.} On a unified-memory device, moving this state into a CPU buffer does
not free capacity, since the CPU and GPU draw from the same memory
pool. An SSD offers a separate storage tier, but saving and restoring
state introduces I/O latency and requires staging buffers in unified
memory. The engine must therefore balance memory used by model
weights and active requests against the state saved for future turns.
Hibana addresses this with SSD-backed session caching and recurrent-state quantization (\S\ref{sec:system}).

%% file: sections/methodology.tex
\section{Methodology}
\label{sec:design}

\label{sec:overview}

Hibana's optimizations fall into three parts, model surgery (\S\ref{sec:model}), unified-memory management (\S\ref{sec:system}), and adaptive speculation (\S\ref{sec:speculation}). We design them together rather than in isolation (\S\ref{sec:speculation} describes where they interact), and Hibana's performance depends on the full set rather than any single component (Figure~\ref{fig:hillclimb}).

\subsection{Model Surgery}
\label{sec:model}
\label{sec:memory}

\subsubsection{Expert Sharing}
\label{sec:moe}

\begin{figure}[t]
    \centering
    \includegraphics[width=0.9\columnwidth]{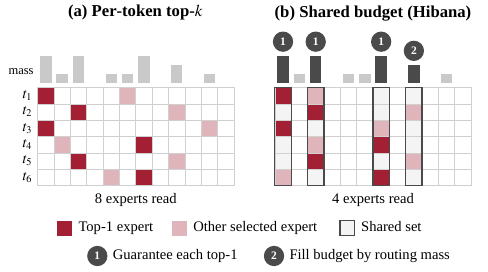}
    \caption{Expert selection in one MoE layer for six tokens with $k=2$.
    Bars show each expert's routing mass summed over the tokens that rank it in
    their top two. (a) Standard routing reads eight expert matrices. (b) Hibana
    guarantees each token's top-1 expert (step 1), fills the rest of a budget
    of four with the expert carrying the most routing mass (step 2), and reads
    four expert matrices.}
    \label{fig:moe-sharing}
\end{figure}

A standard MoE router selects the top-$k$ experts for each token
independently, so a batch of $R$ tokens can read up to $\min(E,kR)$
distinct experts from a layer with $E$ experts, and large batches and
speculative verification read many expert matrices with little reuse.
Hibana instead selects one shared expert set during decode and
verification. Each token is guaranteed its top $g$ experts, and the rest of
the budget goes to the experts with the most routing mass, the router
weight summed over the tokens whose original top-$k$ includes them. Each token then uses up to $k$
of its own candidates that are in the shared set and renormalizes their
weights (Figure~\ref{fig:moe-sharing}), so it can use an expert selected for
another token without adding another expert read. Expert sharing is
separate from the model's always-active shared expert.

For $N$ requests, the budget is
\begin{equation}
 B(N)=\min\!\left(B_{\max},\max(B_{\min},\alpha N)\right),
 \label{eq:expert-budget}
\end{equation}
where $\alpha$ is the budget per request, shared by all of its decode and draft tokens, and $B_{\min}$ and
$B_{\max}$ bound the budget ($\alpha=12$, $B_{\min}=16$, and $B_{\max}=64$
on Flash-Next, where each token routes to $k=10$ of $E=512$ experts).
The budget grows with the number of requests but not with
the verification length, and $g$ shrinks as the batch grows so that the
guarantees alone do not fill the budget. If the guaranteed experts alone
exceed $B(N)$, the layer keeps them, so the budget is a target rather than
a hard cap. 
MoE-Spec~\cite{mcdanel2026moespec} caps the experts each layer loads during speculative verification, and Lynx~\cite{gupta2024lynx} remaps low-affinity token-to-expert assignments within a batch.
Unlike both, expert sharing applies one budget to both decode and verification, scales it with the
number of requests, and always keeps each token's top experts. With $g=2$, this costs 0.8 points of accuracy on average (Table~\ref{tab:expert-ablation}).

\begin{table}[t]
\centering
\caption{Expert-sharing ablation on Qwen3.8-Flash-Next.
$g$ is the number of each token's top experts always kept under a cap of 64
experts per layer. The first row is standard top-10 routing, and the second keeps each token's top five. Accuracy is the paired
change on GSM8K, HumanEval, GPQA, and RULER v2 32K relative to the first
row. Hibana uses $g=2$ at these batch sizes.}
\label{tab:expert-ablation}
\small
\begin{tabular}{lcrrr}
\toprule
 & & \multicolumn{2}{c}{Tokens/s} & Accuracy \\
\cmidrule(lr){3-4}
Selection & $g$ & 16 req. & 64 req. & (pp) \\
\midrule
Standard & 10 & 330 & 702 & -- \\
No cap & 5 & 412 & 765 & $-0.3$ \\
Cap 64 & 3 & 488 & 818 & $-0.5$ \\
\rowcolor{gray!15}
Cap 64 & 2 & 550 & 865 & $-0.8$ \\
Cap 64 & 1 & 597 & 984 & $-1.8$ \\
\bottomrule
\end{tabular}
\end{table}

Table~\ref{tab:expert-ablation} shows how we chose these settings for
Flash-Next. Without a cap, the union of each token's top five experts
reaches 250--460 of the 512 experts per layer at 16--64 requests, so large
batches save little. Under a cap of 64, guaranteeing each token its top
three experts already exceeds the cap, so the cap rarely applies. A top-two
guarantee raises throughput by 23--67\% over standard routing,
and a top-one guarantee is faster still but loses 1.8 points on average. The
minimum budget $B_{\min}=16$ comes from a single request. With 12 experts,
GSM8K fell by 4.7 points, and with 16 by 0.3 points, within noise.
Appendix~\ref{sec:expert-routing-details} gives model-specific details.

Expert sharing also lowers the marginal cost of verifying more draft
tokens, because they can reuse experts already read for the batch. This is codesigned with the adaptive speculation
policies in \S\ref{sec:adaptive}.

\subsubsection{Two-Bit Draft Head}
\label{sec:draftcomp}

Hibana drafts with the model's own multi-token prediction (MTP)
head~\cite{gloeckle2024multitoken}, a small module trained with the target
that predicts the next tokens autoregressively from the target's hidden
state, or, for Qwen3.8-27B, with DFlash2~\cite{inco2026dflash2}, a diffusion
draft model that proposes a block of tokens in one parallel step.
Each MTP draft step projects a hidden state onto a vocabulary of about
249k tokens, and reading the FP8 output head for this projection costs
about 640\,MB per step in terms of memory footprint. 
Hibana instead ranks every token with a two-bit
copy of the head (171\,MiB), keeps the 16 highest-scoring candidates, and
rescores only those with the matching rows of the target's FP8 head, about
40\,KB. The draft emits the candidate with the best FP8 score.
The two-bit copy uses the codebook format and kernels that Hibana uses for its
three-bit weights (\S\ref{sec:weights}) with four codebook entries instead
of eight, the FP8 Lloyd-Max levels of a Gaussian, so each weight costs 2.25
bits including its group scale. Because four levels leave little range, each
group's scale is chosen from a few fractions of the group's maximum by least
squared error rather than set to the maximum itself.

\begin{figure}[t]
\centering
\includegraphics[width=\linewidth]{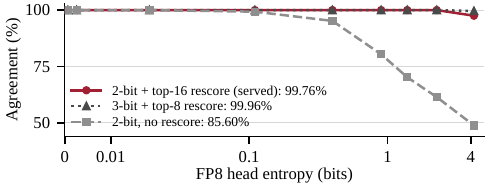}
\caption{Agreement with the FP8 head's token by the FP8 head's entropy,
over 24,460 draft steps of Qwen3.8-Flash-Next. Legend values are overall
agreement.}
\label{fig:draft-head}
\end{figure}

The majority of the computation runs in two bits, while the final choice uses FP8 scores.
Two-bit quantization perturbs each logit slightly, which changes the winner only when the top tokens' scores are very close. 
These steps benefit from the FP8 head, in which the top token almost always remains among the 16 candidates.
Figure~\ref{fig:draft-head} confirms this.

Without rescoring, agreement with the FP8 head falls as the FP8 head's entropy rises, to below 50\% on
the least certain steps and 85.6\% overall. With rescoring, it stays near
100\% at every entropy level and reaches 99.8\% overall.
The target still verifies with its own head, so this approximation changes
only which tokens are proposed, not which are accepted. Replaying the
recorded rounds yields an unchanged 3.532 accepted tokens per round with the two-bit head and the full draft step of Qwen3.8-Flash-Next at batch one on one DGX Spark decreases from 1.87\,ms to 1.55\,ms
(Appendix~\ref{sec:draft-head-details}).

\subsubsection{Mixed-Precision Weights}
\label{sec:weights}

Hibana stores routed experts, and a decode copy of each dense projection,
as three-bit indices into an eight-entry FP8 (E4M3) codebook. Each group of
32 weights has an E4M3 scale and each output row an FP32 scale, so the
indices and group scales take 3.25 bits per weight, 72\% of NVFP4's 4.5.
Hibana fits the indices and scales to calibration activations with
GPTQ-style updates~\cite{frantar2023gptq}, then recalibrates the dense decode copies against the original model to preserve quality.
Codebook formats for LLM weights
have been studied before~\cite{park2022lutgemm,tseng2024qtip,guo2024flute},
and Hibana executes this format on FP8 tensor cores. Its kernels store the
packed bits in the order tensor-core instructions consume them, expand them
in registers into FP8 operands without a separate dequantization pass, and
multiply them by FP8 activations with FP32 accumulation. 

The precision depends on the layer and the batch. Three-bit experts replace
the NVFP4 copies at every batch size, prefill included, which reduces both memory and weight
traffic. Dense projections keep their FP8 or NVFP4 matrices for prefill and
add a three-bit copy, at some memory cost, for calls of at most 128 tokens such as decode and verification.
Table~\ref{tab:w3-microbenchmarks} compares our three-bit MoE kernel with
FlashInfer's CUTLASS NVFP4 fused-MoE kernel. It is 1.20--1.36$\times$ faster from 512 to 2k
tokens, close to the 1.38$\times$ ratio of their bits per weight, and
1.06$\times$ faster at our 4k-token prefill chunk.

\begin{table}[h]
\centering
\caption{Latency of one Flash-Next MoE layer, Hibana's three-bit kernel vs.\ FlashInfer's CUTLASS NVFP4 fused-MoE kernel.}
\label{tab:w3-microbenchmarks}
\small
\setlength{\tabcolsep}{4pt}
\begin{tabular}{rrrr}
\toprule
Tokens & Three-bit (ms) & CUTLASS NVFP4 (ms) & Speedup \\
\midrule
512 & 4.90 & 6.68 & $1.36\times$ \\
1k & 5.53 & 7.22 & $1.30\times$ \\
2k & 6.84 & 8.24 & $1.20\times$ \\
4k & 9.90 & 10.53 & $1.06\times$ \\
\bottomrule
\end{tabular}
\end{table}

\subsection{Unified Memory}
\label{sec:system}

On a unified-memory device, weights, active requests, saved sessions, and
I/O buffers draw on the same memory, and the SSD is the only separate
capacity tier. Hibana keeps inactive sessions on the SSD through a small
fixed staging buffer, stores recurrent state in INT8 so that every
session takes less memory, and serves large Engram embedding
tables~\cite{cheng2026conditionalmemoryscalablelookup} from the SSD.

\subsubsection{SSD Session Cache}
\label{sec:context}
\label{sec:hicache}

To resume an inactive agent, the inference engine needs its KV pages, its recurrent state, and the draft model's state; we call this bundle a \emph{session}.
Inactive agent states occupy capacity that active requests need, yet evicting them forces repeated prefill, which is costly especially on local devices. 
Hibana stores its prefix cache in unified memory (L1) and uses the SSD as an L2 cache.
We call the cache that keeps sessions on the SSD the \emph{SSD session cache}.

\begin{figure}[t]
    \centering
    \includegraphics[width=0.85\columnwidth]{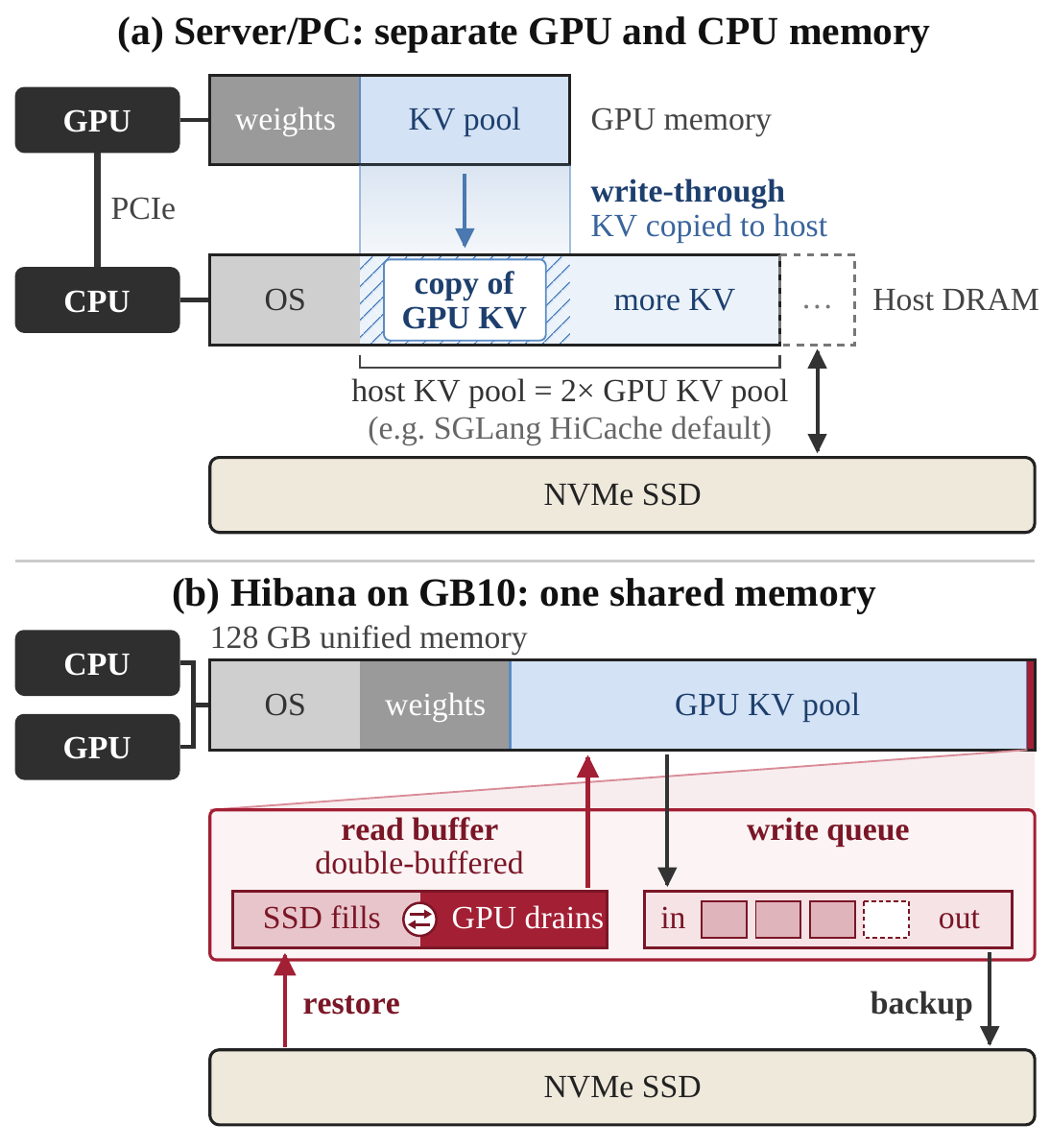}
    \caption{KV-cache tiers on separate versus unified memory. (a) On a
    server or PC, host DRAM is separate memory, so a host tier that keeps a
    copy of the GPU's KV adds capacity. (b) On GB10 the CPU and GPU share one
    memory. Hibana keeps no host copy and moves KV between the GPU's KV pool
    and the SSD through about 1\,GiB of staging, the same at any context
    length: restores stream through a double-buffered read buffer, and backups
    wait in a write queue. Only the staging slice is drawn to scale.}
    \label{fig:ssd-staging}
\end{figure}

On a server or PC, the GPU has its own memory, and host DRAM is a
separate, usually larger pool. Tiered KV caches therefore use host DRAM as
a second tier that holds a copy of the GPU's KV cache; SGLang's
HiCache, for example, sizes it at twice the GPU pool by
default~\cite{sglang2026hicache} (Figure~\ref{fig:ssd-staging}a). 
On GB10, the CPU and GPU share one 128\,GB memory, so Hibana uses the SSD directly as the second tier session cache. 

Hibana cannot read the SSD straight into the KV pool. DGX Spark lacks
GPUDirect Storage, so the drive transfers data only to and from host
memory, and a stored block is one contiguous run of bytes while its KV is
spread over many per-layer pages. Every transfer therefore goes through a
staging buffer in host memory, and a GPU kernel moves the data between
that buffer and the KV pages.

The staging buffer comes out of the same memory as the KV cache, so Hibana keeps it small and fixed: about 1\,GiB, the same at any context length (Figure~\ref{fig:ssd-staging}b). 
The buffer is pinned, so its pages stay in place while the drive and the GPU use it, and all transfers use direct I/O, which bypasses the operating system's file cache instead of keeping a second copy of the data in memory.
The buffer is much smaller than a long prefix, so a restore reads the prefix in chunks. 
With one buffer, the SSD and the GPU would take turns: the drive could not load the next chunk until the GPU had copied the current one into KV pages.
Hibana splits the read buffer into two halves (Figure~\ref{fig:ssd-staging}b): while the SSD fills one half, the GPU drains the other, and then they swap.
The GPU drains a half much faster than the SSD fills one, so the drive rarely waits and a restore runs at close to the drive's read bandwidth. A longer prefix takes more rounds through the same buffer, not a larger buffer.
On GB10 the GPU reads the buffer in place, so draining it needs no extra copy into GPU memory.
SSD writes are slow, so a GPU kernel first copies KV pages into a write queue, allowing their GPU memory to be reused immediately. Writer threads then save the copies to SSD in the background. The queue has a fixed size to limit memory use. If it is full, evicted pages are discarded without a backup, so requests never wait for SSD writes.

\subsubsection{Recurrent-State Quantization}
\label{sec:statecompression}

Each recurrent layer of a hybrid model summarizes all earlier tokens in a
fixed-size state matrix, which every decode step reads and updates.
Hibana keeps a snapshot of this state for every saved session, for every
subagent that branches from a shared prefix, and for every speculative
candidate under verification, so at full precision these snapshots grow
with the number of agents. Hibana therefore stores recurrent state in
INT8 using LeapQuant~\cite{pan2026leapquant}, which builds on earlier
quantization of state-space models~\cite{chiang2024quamba,chiang2025quamba2};
Nemotron~3 Ultra likewise caches its recurrent state in
INT8~\cite{nvidia2026nemotron3ultra}. Quantizing the state after every
token would compound rounding error, so LeapQuant quantizes it only once per
window of tokens. At the start of a window, the state is stored as an INT8
matrix with its scales, and its largest outlier entries are kept separately
at high precision as a few compensator tokens that pass through the same
update path as real tokens. Within the window, each new token's update is
appended to a high-precision buffer instead of being written into the INT8
matrix, and outputs are computed from the INT8 matrix together with the
buffered updates. At the end of the window, the buffered updates are folded
into the state, which is quantized again, so rounding error enters once per
window rather than once per token.

The same buffer makes verification and session saves inexpensive. During
verification, the updates of candidate tokens are held in a temporary
buffer, and committing a round appends only the updates of the accepted
tokens, so Hibana never copies a full recurrent snapshot per candidate and
rejected tokens never reach the committed state. A saved session stores the
INT8 matrix, its compensator tokens, and the buffered updates together,
which also reduces the bytes the SSD session cache writes and restores.

\subsubsection{Engram Offloading}
\label{sec:ple}

Engram tables use the SSD for sparse model-data reads during execution.
Hibana keeps the table on the SSD and memory-maps it, so the operating system's page cache holds recently used rows. Each lookup first copies the rows already in the page cache without blocking, then reads the missing rows in parallel from a thread pool, so their SSD latencies overlap instead of adding up.
Prefetch overlaps future prompt-chunk or already-proposed-token lookups
with available GPU work. Cached rows still consume unified memory, but the
whole table need not remain resident; configurations with sufficient
capacity can retain their partition in memory. This lookup path is separate
from restoring a session and is governed by row locality and the
compute available to overlap I/O.

\subsection{Adaptive Speculation}
\label{sec:speculation}

Hibana uses one execution lifecycle for multi-token prediction (MTP)
heads, autoregressive draft models, and diffusion draft models such as
DFlash2~\cite{inco2026dflash2}, which reserves candidate state, drafts,
verifies, accepts, commits, and feeds the accepted result back to the draft
model. Around this lifecycle, Hibana adapts
how much to draft and verify to the expected acceptance
(\S\ref{sec:adaptive}), accounting for how expert sharing
(\S\ref{sec:moe}) lowers the cost of verification on MoE targets, and
optionally relaxes the acceptance rule (\S\ref{sec:relaxed}).

\subsubsection{Adaptive Drafting and Verification}
\label{sec:adaptive}

The objective is to maximize the expected number of emitted tokens per
unit of time, including drafting, verification, and commit overhead. An
additional draft token is useful only when its chance of being accepted
repays the time it takes to draft and verify. Hibana controls this at two
points in a round: \emph{adaptive drafting} chooses how many tokens an
autoregressive (AR) draft model such as MTP drafts, and \emph{adaptive
verification} chooses how many tokens of an already-drafted diffusion block
the target verifies.

\paragraph{Adaptive drafting for AR draft models.}
Prior work adapts the draft length with a learned stopping
predictor~\cite{huang2024specdecpp} or with dynamic draft
trees~\cite{li2024eagle2}. Choosing the draft length before drafting avoids
both the draft steps and their verification when acceptance is unlikely.
For MTP, Hibana maintains a per-request acceptance rate $q$, an
exponentially smoothed ratio of accepted draft tokens to tested positions,
where a rejected round counts the accepted tokens and the first rejected
position as tested. The desired draft length is
\begin{equation}
 d^\star = \arg\max_{1\leq d\leq d_{\max}}
 \frac{\sum_{j=0}^{d}q^j}{1+c(d-1)},
 \label{eq:adaptive-depth}
\end{equation}
where $d_{\max}$ is a cap that depends on the batch size. The numerator is
the expected number of emitted tokens, including the target's additional
token, assuming each position is accepted independently with probability
$q$. The denominator is the round time relative to a round with one draft
token, where $c$ is the time that one more draft step and its verification
add, as a fraction of that round's time. $c$ is a fixed value configured
per model, not measured online.

Hibana computes each request's desired draft length before the round from
its completed rounds. Each round runs as a captured CUDA graph, one per
combination of batch size and draft length. Small batches use the longest
desired draft length among their requests, up to the cap. Larger batches
use a fixed draft length per range of batch sizes, which avoids capturing a
graph for every draft length at every batch size. All requests in a batch
therefore draft the same number of tokens. The caps also bound temporary
state and account for weight-dispatch and expert-reuse effects that the
simple cost model omits.

\paragraph{Adaptive verification for diffusion drafters.}
For a diffusion drafter, Hibana adapts how many drafted tokens the target
verifies rather than how many it drafts. DFlash2 drafts a whole block in one
parallel step, so drafting fewer tokens saves little, while verifying fewer
saves target work. EVICT~\cite{pan2026adaptiveverify} also truncates drafts
before verification, aimed at MoE targets where each additional verified
token can activate more experts. Hibana instead truncates only on the dense
Qwen3.8-27B target, because on MoE targets expert sharing already bounds the
experts that verification reads (\S\ref{sec:moe}). For Qwen3.8-27B, the policy predicts how
likely each draft token is to be accepted from the drafter's confidence
scores and adjusts these predictions with the acceptance observed at each
position. A later token counts only if every token before it is accepted,
so its expected contribution includes that probability. Using measured
latencies of the CUDA graphs that verify each prefix length and an estimate
of the remaining round overhead, the policy selects the prefix length for
each request that maximizes expected emitted tokens per unit of time.
Requests with different prefix lengths are packed into one verification
pass, so a confident request verifies more tokens than an uncertain one.
Periodic rounds that verify the full block keep the estimates for later
positions fresh.

This policy applies to greedy batches without grammar constraints. Sampled
batches choose a shared verification length before drafting, so the choice
does not depend on the sampled draft tokens, and greedy batches containing
a grammar-constrained request verify the configured maximum length.
Choosing how much to verify is separate from deciding which draft tokens to
accept (\S\ref{sec:relaxed}).

\begin{table}[t]
\centering
\caption{Decode speedup of adaptive drafting over a fixed draft length
(MTP on Qwen3.8-Flash-Next) and of adaptive verification over verifying
the full draft (DFlash2 on Qwen3.8-27B).}
\label{tab:adaptive}
\small
\begin{tabular}{lrr}
\toprule
Policy & Requests & Speedup \\
\midrule
Adaptive drafting & 1 & $1.16\times$ \\
 & 4 & $1.05\times$ \\
\midrule
Adaptive verification & 8 & $1.13\times$ \\
 & 16 & $1.28\times$ \\
\bottomrule
\end{tabular}
\end{table}

Table~\ref{tab:adaptive} shows that adaptive drafting helps most at one
request, where the request sets its own draft length, while adaptive
verification helps most in larger batches, where verifying every request's
full draft is costly.

\paragraph{Adaptive speculation under expert sharing.}
On MoE targets, expert sharing (\S\ref{sec:moe}) bounds the number of
experts a verification pass reads, so verifying more draft tokens adds
little cost, and truncating a draft saves little time while discarding
tokens that might have been accepted. Hibana therefore verifies
Flash-Next's diffusion drafts at a tuned length without truncation, and for
Nemotron's MTP head tunes the expert budget together with the draft length,
because verifying too many tokens under a fixed budget weakens each token's
expert coverage.

\subsubsection{Relaxed Verification}
\label{sec:relaxed}

Exact speculative sampling can reject a draft token even when the target
considers it plausible, reducing the number of tokens produced per round.
Medusa's typical acceptance~\cite{cai2024medusa}, Judge
Decoding~\cite{bachmann2025judge}, and loosely speculative
decoding~\cite{li2025fly} accept such drafts beyond an exact match.
We introduce a relaxation parameter $s\geq 0$ to control this tradeoff.
Let $p$ and $q$ denote the target and draft sampling distributions at the
current history. For an eligible draft token $y\sim q$, the acceptance
probability is $a_s(y)=\min\left(1, e^s p(y)/q(y)\right)$.
Setting $s=0$ recovers exact speculative sampling. Increasing $s$ accepts
more proposals, allowing more accepted tokens per round while changing the output
distribution. Relaxation is restricted to tokens within the target's
top-$k$ predictions and a configured log-probability margin of its most
likely token; other proposals retain the exact acceptance rule. On
rejection, we retain the standard residual sampling procedure.
A finite $s$ keeps a probability-based check against the target: passing
the rank and margin thresholds alone does not guarantee acceptance.
As $s\to\infty$, eligible proposals with positive target probability
are always accepted. At a fixed history, the resulting next-token
distribution $P_s$ satisfies $P_s(x)\leq e^s p(x)$
(Appendix~\ref{sec:relaxed-bound}); Appendix~\ref{sec:relaxed-sweep}
sweeps $s$. The fast profile uses $s=0.5$ and treats a draft as
eligible when it is among the target's top two tokens and within 1.0 logit
of its top token. 
Table~\ref{tab:ablation} reports the accuracy impact of relaxed verification in the evaluated fast profile, and Figure~\ref{fig:hillclimb} shows its incremental contribution to decoding speedup.
On Qwen3.8-Flash-Next with its MTP drafter, relaxed verification raises the accepted tokens per verify step from 2.15--2.34 under exact verification to 2.76--3.35 at $s=0.5$, 26--43\% more, with no clear loss in accuracy (Appendix~\ref{sec:relaxed-sweep}).

%% file: sections/evaluation.tex
\section{Evaluation}
\label{sec:evaluation}

\subsection{Experimental Setup}

\paragraph{Hardware and targets.}
All experiments run on one DGX Spark, whose GB10 CPU and GPU share 128\,GB
of memory~\cite{nvidia2026sparkhardware}.
Appendix~\ref{sec:two-sparks} reports Hibana on two DGX Sparks. We evaluate four targets, the
dense Qwen3.8-27B~\cite{qwen2026qwen38_27b} and three MoE models,
Qwen3.8-Flash-Next~\cite{qwen2026qwen38flashnext}, Nemotron-3.5-Bolt~\cite{nvidia2026nemotron35bolt}, and
Nemotron-3.5-Super~\cite{nvidia2026nemotron35super}. Hibana runs its fast
profile, which enables all of its optimizations, unless stated otherwise, and its exact profile serves the original model with exact verification.

\paragraph{Baselines.}
We compare against vLLM~\cite{kwon2023pagedattention}, llama.cpp~\cite{llamacpp2026}, Flare~\cite{nvidia2026flare}, and
TensorFold~\cite{tensorfold2026}. Where supported, each baseline runs NVFP4 weights with the fastest drafter we could run for the target, DFlash2~\cite{inco2026dflash2} for Qwen3.8-27B and the model's MTP head for the others, unless stated otherwise. llama.cpp runs its own quantized weights.

\paragraph{Benchmarks.}
We measure decoding throughput on SPEED-Bench~\cite{abramovich2026speedbench}
with reasoning enabled. We measure end-to-end agentic performance by
replaying recorded agent sessions from
AgentPerf-Local~\cite{artificialanalysis2026agentperflocal}, and we open more agent sessions than
GPU memory can hold to stress the cache (\S\ref{sec:eval-agents}). We measure
task accuracy on GPQA~\cite{rein2023gpqa},
LiveCodeBench (LCB)~\cite{jain2024livecodebench},
IFBench~\cite{pyatkin2025ifbench}, and AIME 2025 and 2026~\cite{aime26}.

\subsection{Decoding Throughput}
\label{sec:eval-e2e}

\begin{figure*}[t]
    \centering
    \includegraphics[width=1.0\linewidth]{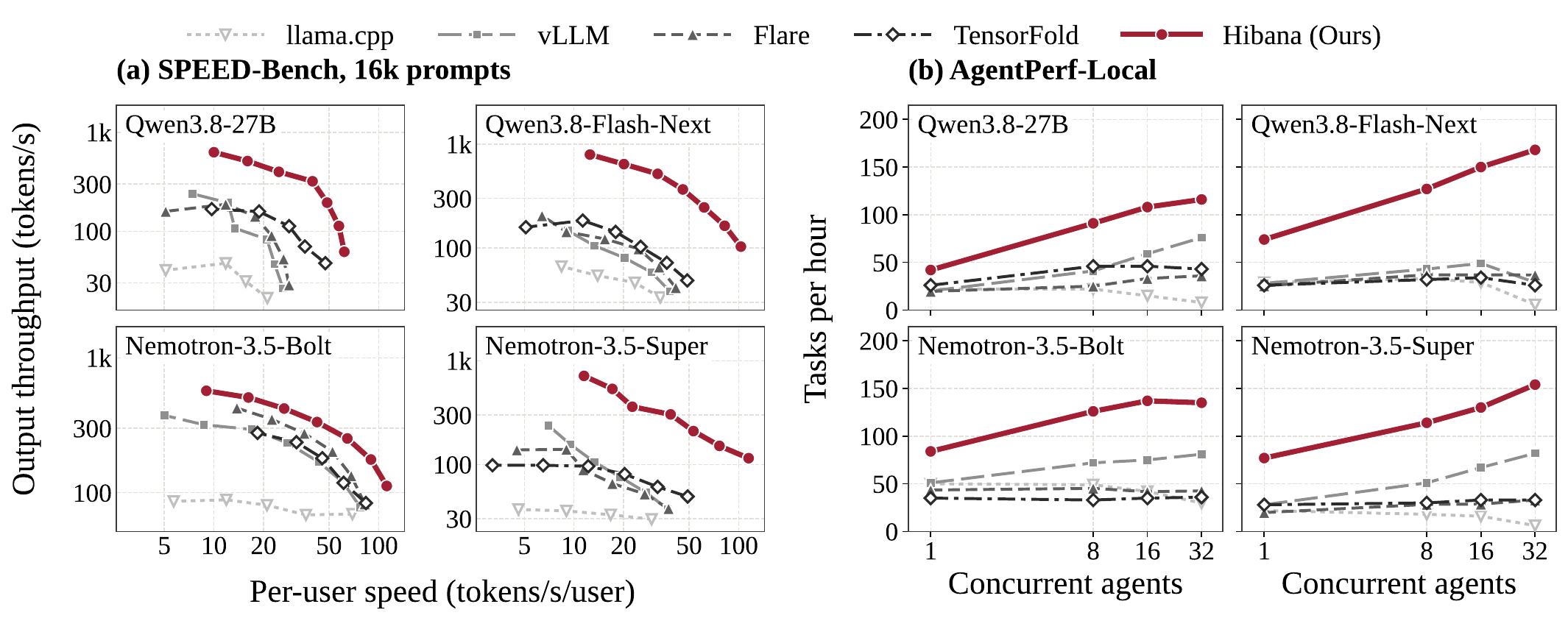}
    \caption{(a) SPEED-Bench throughput on one DGX Spark as concurrency grows.
    Each point is one concurrency level, labeled with its number of
    requests. Up and to the right is better. (b) Agentic task throughput on AgentPerf-Local replay as the number of concurrent agents grows. Missing points mark configurations
    the baseline could not run.}
    \label{fig:speed-eval}
\end{figure*}

Figure~\ref{fig:speed-eval} sweeps concurrency on SPEED-Bench. On all four
targets, Hibana reaches higher throughput than vLLM, Flare, TensorFold, and
llama.cpp at every concurrency level. Hibana's three-bit weights and
compact KV cache leave memory for up to 64 concurrent requests; we run each
baseline up to the most it can serve, at most 32.
The gain is largest on Qwen3.8-Flash-Next,
where Hibana's peak throughput (at 64 requests) is $3.9\times$ that of
the strongest baseline's peak (Flare at 32 requests). At the same 32
requests the gap is $3.1\times$.
\subsection{Agentic Tasks}
\label{sec:eval-agents}

\paragraph{Task throughput.}
Figure~\ref{fig:speed-eval}b reports end-to-end agentic performance. At each
system's best concurrency, Hibana completes 1.5--3.4$\times$ more tasks
per hour than the fastest baseline. Qwen3.8-Flash-Next gains the most:
Hibana keeps rising to 168 tasks per hour at 32 agents, while every
baseline stalls or slows past 16.

\paragraph{Task throughput needs both decoding speed and cache reuse.}
TensorFold decodes faster than vLLM at low concurrency on most targets
(Figure~\ref{fig:speed-eval}a), yet from 16 agents on, vLLM completes more
tasks per hour on every target (Figure~\ref{fig:speed-eval}b). Some
AgentPerf-Local traces rewrite the end of the previous prompt every
turn. TensorFold reuses a cached prompt only when the new one extends it
exactly, so on Qwen3.8-Flash-Next it serves 0\% of these prompts from
cache, against 44\% for vLLM, which resumes from its last saved point;
TensorFold also serves at most 16 streams. Flare
would resume from the right point, but its checkpoint pool is fixed and
shared. Each request keeps only five checkpoints on Qwen3.8-Flash-Next,
so the one before the rewritten end is usually missing, and more agents
evict each other's. Its hit rate is 26--56\% at one agent and 19\% at 32
(Appendix~\ref{sec:agent-reuse}).
Hibana leads on both factors:
it decodes fastest at every concurrency (Figure~\ref{fig:speed-eval}a) and
keeps 93--96\% of these prompts in cache. With the rewrite removed, so all
engines reuse context equally well, Hibana still completes about
$2\times$ as many tasks per hour as vLLM, which isolates its decoding
advantage (Appendix~\ref{sec:agent-reuse}). On Qwen3.8-Flash-Next, Hibana's lead at 16 agents falls from $3.1\times$ to $1.9\times$ without the turn note.

\paragraph{When sessions outgrow GPU memory.}
\begin{figure}[t]
    \centering
    \includegraphics[width=0.85\linewidth]{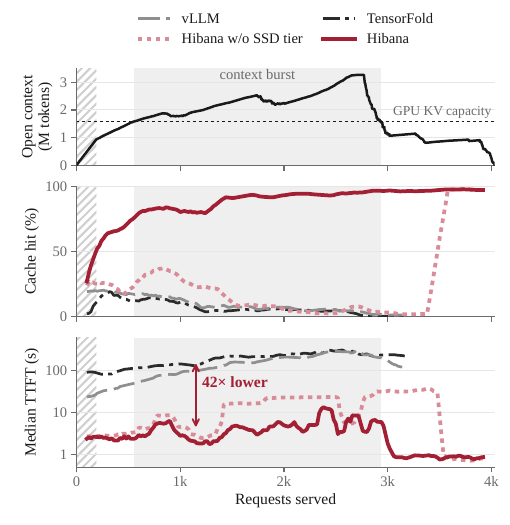}
    \caption{Qwen3.8-Flash-Next serving 192 agent conversations whose
    combined context reaches twice GPU KV capacity. Top: live session
    context, the tokens of all unfinished conversations. Middle: share of
    each prompt served from cache. Bottom: median TTFT, log scale. 32 agents
    send turns back to back, with thinking off and no tool time. Curves are
    200-turn rolling medians; the hatched region is warm-up. Baseline runs
    were stopped after about 3.3k turns.}
    \label{fig:burst}
\end{figure}

Figure~\ref{fig:burst} shows that when agent sessions hold more
context than GPU memory, only Hibana's SSD tier keeps them cached. We
replay 192 AgentPerf-Local conversations at once, so their combined
context grows to about twice Hibana's GPU KV capacity.
Caches fail here because the conversations take turns: about 130 turns
pass before the same conversation returns. If GPU memory cannot hold them
all, each one is evicted before it returns, and its next turn
recomputes most of its context.
On these hybrid models they can keep the saved state of only 16 to
32 conversations, so their hit rates stay below 20\%. For vLLM, the
limit comes from its page layout: each saved recurrent state takes a
KV-cache page enlarged to hold about 3,000 tokens, so few states fit, and
a cached prefix is reused only in whole blocks of that size.
Hibana without the
SSD tier holds more, but its hit rate stays below about 35\%, and once
live session context passes about 2\,M tokens it falls below 10\% and each
turn recomputes about 11,000 tokens.

The SSD tier removes this limit. Hibana writes each evicted conversation
to the SSD and reads it back on the conversation's next turn
(\S\ref{sec:context}). Its hit rate climbs to 80--98\%, and each turn
recomputes about 600 tokens instead of 11,000. As a result, its median
TTFT stays at a few seconds, while it rises to
15--35\,s without SSD offloading. The baselines' TTFT exceeds 100\,s, up to $42\times$ Hibana's. This larger gap also comes from Hibana's higher throughput, so each turn waits less behind other agents' prefill and decoding. Even when every engine misses the cache, Hibana without the SSD tier has a median TTFT of 25\,s, against 210--245\,s for the baselines.

\subsection{Where the Speedup Comes From}
\label{sec:eval-ablation}

\begin{figure}[t]
\centering
\includegraphics[width=0.8\linewidth]{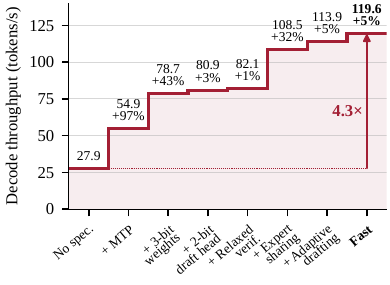}
\caption{Decode throughput on Qwen3.8-Flash-Next at one request as each
optimization is added. Percentages are the change from the previous step.
The baseline runs NVFP4 experts without speculation. The last step adds an
NVFP4 KV cache.}
\label{fig:hillclimb}
\end{figure}

Figure~\ref{fig:hillclimb} adds Hibana's optimizations one at a time on
Qwen3.8-Flash-Next, from a baseline without speculation to the fast
profile. Three steps account for most of the gain. MTP drafting with a draft length of 4
raises throughput by 97\%, three-bit weights add 43\%, and expert sharing
adds a further 32\%. Adaptive drafting and the final step add 5\% each, and
the two-bit draft head and relaxed verification add 3\% and 1\%. In total,
throughput rises from 27.9 to 119.6 tokens/s, a $4.3\times$ speedup.

\subsection{Accuracy of the Lossy Optimizations}
\label{sec:eval-quality}

\begin{table}[h]
\centering
\caption{Task accuracy (\%) on Qwen3.8-Flash-Next as each lossy
optimization is added. The last row is the full fast recipe.
Parentheses give the change from the exact baseline.}
\label{tab:ablation}
\small
\setlength{\tabcolsep}{3pt}
\newcommand{\dn}[1]{\,{\scriptsize($-$#1)}}
\resizebox{\columnwidth}{!}{\begin{tabular}{lrrrrr}
\toprule
 & GPQA & LCB & IFBench & AIME25 & AIME26 \\
\midrule
Exact               & 90.7 & 87.7 & 85.3 & 100.0 & 100.0 \\
+ 3-bit weights     & 90.4\dn{0.3} & 85.4\dn{2.3} & 83.3\dn{2.0} & 99.2\dn{0.8} & 99.2\dn{0.8} \\
+ Expert sharing    & 90.4\dn{0.3} & 85.4\dn{2.3} & 83.3\dn{2.0} & 99.2\dn{0.8} & 99.2\dn{0.8} \\
+ Relaxed verif.    & 89.1\dn{1.6} & 84.6\dn{3.1} & 81.7\dn{3.6} & 99.2\dn{0.8} & 99.2\dn{0.8} \\
\bottomrule
\end{tabular}}
\end{table}

Exact speculative sampling preserves a fixed target's output
distribution~\cite{leviathan2023speculative}, but three-bit weights, expert
sharing, and relaxed verification can change it. Table~\ref{tab:ablation}
adds these optimizations one at a time on Qwen3.8-Flash-Next. Three-bit
weights cost 2.3 points on LCB, 2.0 on IFBench, 0.8 on AIME, and 0.3 on
GPQA.
Expert sharing leaves every score unchanged.
Relaxed verification with $s=0.5$ costs a further 1.3 points on GPQA,
0.8 on LCB, and 1.6 on IFBench, and leaves AIME26 unchanged. The last row is the complete fast
recipe, which also enables the NVFP4 KV cache. It stays within 3.6 points of the exact baseline on every
benchmark it was run on. Appendix~\ref{sec:accuracy-other-targets} reports the same
comparison on Nemotron-3.5-Super and Qwen3.8-27B.

%% file: sections/related_work.tex
\section{Related Work}
\label{sec:related}

\paragraph{Datacenter serving.}
vLLM manages the KV cache in fixed-size pages so that many requests can be
batched with little wasted memory~\cite{kwon2023pagedattention}, and SGLang
reuses shared prefixes across requests and, with HiCache, extends this reuse
to host memory and storage~\cite{zheng2024sglang,sglang2026hicache}. Both
target discrete datacenter GPUs, whereas Hibana applies batching and prefix
reuse within the shared memory and limited bandwidth of one unified-memory
device.

\paragraph{Local serving.}
llama.cpp and MLX LM run quantized models on consumer
hardware~\cite{llamacpp2026,hannun2023mlx,mlxlm2026}, and llama.cpp's
server adds continuous batching, tool calling, and speculative
decoding~\cite{llamacpp2026server}. NVIDIA Flare targets long conversations
and branching agents on DGX Spark and checkpoints attention, recurrent, and
speculative state~\cite{nvidia2026flare}, and TensorFold serves concurrent
requests with disk-backed conversation and prefix reuse on
Macs~\cite{tensorfold2026}. Hibana shares their goal of local agent serving
but also changes the model and the decoding procedure, so that expert
sharing, compressed state, and adaptive speculation reduce the memory
traffic behind each accepted token, which limits the benefit of speculative
decoding on these devices~\cite{liu2026speculative}.

\paragraph{Memory-constrained execution.}
FlexGen places and compresses weights across GPU memory, host memory, and
disk for batched inference~\cite{sheng2023flexgen}, and LLM in a flash
loads parameters from flash storage on demand~\cite{alizadeh2024flash}.
FreeToken and Strata bring expert placement, agent-state reuse, and session
save and restore to edge MoE
serving~\cite{yang2026freetoken,freetoken2026runtime,strata2026}. Hibana
instead keeps execution in unified memory, compresses weights and state to
leave room for active sessions, and places only inactive sessions and
Engram tables on SSD. Tutti restores SSD-backed KV through GPU-initiated
I/O~\cite{qiu2026tutti}. Like Flare and Strata, Hibana saves complete
sessions, including attention, recurrent, and draft state, but stores
recurrent state in INT8 and restores through a small fixed staging buffer,
so persistence takes little memory from active requests.

%% file: sections/conclusion.tex
\vspace{-2pt}
\section{Conclusion}
\label{sec:conclusion}

We presented Hibana, an inference engine for agentic workloads on
unified-memory devices. Hibana combines model surgery, unified-memory
management, and adaptive speculation, and designs them together around the shared memory capacity and limited bandwidth of a
unified-memory system-on-chip. On a single DGX Spark serving Qwen and
Nemotron models, Hibana completes up to $3.4\times$ more agentic tasks per
hour and delivers up to $3.9\times$ higher decoding throughput than vLLM,
Flare, llama.cpp, and TensorFold.

%% file: sections/AI_usage_statement.tex
\section*{AI Usage Disclosure}

The authors utilized ChatGPT (OpenAI) and Claude (Anthropic) as
supporting tools throughout the research and manuscript preparation
process.

During the research phase, these tools facilitated brainstorming,
exploration of methodological approaches and mathematical
formulations, and identification, synthesis, and comparison of
relevant literature. During implementation and evaluation, they
provided assistance with code development, refinement, and debugging,
experimental design, selection of baselines, and interpretation
of experimental findings.

For manuscript preparation, the tools assisted with drafting and
editing, improving language clarity and grammatical accuracy,
developing and refining figures and tables, and reviewing
submission guidelines, formatting, and references.

All research decisions, findings, and submitted content were
reviewed and validated by the authors, who assume full
responsibility for the accuracy, originality, and integrity
of the work.

%% file: sections/appendix.tex
\section{Expert-Sharing Implementation Details}
\label{sec:expert-routing-details}

The expert sharing procedure in \S\ref{sec:moe} preserves each model's
router ranking and weighting conventions. Flash-Next scores an expert by
summing its softmax probability over tokens whose original top-$k$ includes
it. Nemotron ranks candidates using sigmoid scores plus its expert-selection
bias, but aggregates the unbiased sigmoid weights normalized over each
token's original top-$k$. After constructing the shared set, Flash-Next
selects up to $k$ eligible experts from a wider ranked candidate list;
Nemotron retains eligible experts from its original top-$k$.

In the request-scaled configuration, Flash-Next uses a minimum budget of
16 experts; Nemotron has no additional minimum. For example,
$\alpha=12$, $B_{\min}=16$, and $B_{\max}=64$ yield budgets of 16, 24,
48, and 64 experts for 1, 2, 4, and 8 real requests. These are configured
budgets, not measured expert counts. Flash-Next decreases the per-token
guarantee as the execution shape grows. Nemotron applies a smaller fallback
guarantee if the initial guaranteed union exceeds the budget. Both preserve
the resulting guaranteed set even if it exceeds the cap. Guarantee sizes
are derived from the execution shape, including graph-bucket dimensions;
the shared-set budget counts only real requests.

\section{Two-Bit Draft Head Agreement}
\label{sec:draft-head-details}

Figure~\ref{fig:draft-head} compares the two-bit draft head with and
without FP8 rescoring (\S\ref{sec:draftcomp}) on 24,460 recorded greedy draft steps of Qwen3.8-Flash-Next in the
fast profile. Each step is binned by the entropy of the FP8 head's output
distribution, and each curve reports how often it selects the same
token as the full FP8 head.

On confident steps, below about 0.1 bits of entropy, both variants
agree with the FP8 head, because the top token leads by a margin larger
than the two-bit quantization error. As entropy rises, the leading tokens
score more closely and the two-bit head alone increasingly picks a
different winner, falling to about 80\% agreement near one bit and below
50\% at four bits. Rescoring removes this loss. The two-bit ranking still
places the FP8 winner among its 16 highest-scoring candidates, so
recomputing those 16 scores with FP8 rows recovers it. The served two-bit
head with 16 candidates agrees on 99.76\% of steps overall, close to the
earlier three-bit head with 8 candidates at 99.96\%. Replaying the
recorded rounds yields 3.532 accepted tokens per round for both.

The rescoring pass reads only 16 rows of the FP8 head, under 0.01\% of
the full head. On one DGX Spark, the head projection decreases from
1.11\,ms with the three-bit head to 0.78\,ms with the two-bit head and
rescoring, which accounts for the 1.87 to 1.55\,ms reduction in the full
draft step reported in \S\ref{sec:draftcomp}.

\section{Relaxed Verification}
\label{sec:soft-acceptance-bound}

Relaxed verification (\S\ref{sec:relaxed}) accepts an eligible draft
token $y$ with probability $\min(1, e^{s}p(y)/q(y))$, where $s\geq0$ is the
relaxation parameter and $s=0$ recovers exact speculative sampling. This
appendix bounds how much this rule can amplify any token's probability
(\S\ref{sec:relaxed-bound}) and measures how accuracy and
acceptance change with $s$ (\S\ref{sec:relaxed-sweep}).

\subsection{Output Distribution}
\label{sec:relaxed-bound}

Fix a decoding history and let $p$ and $q$ be the normalized target and
draft distributions used by the sampling procedure, including its
temperature and filtering settings. Let $S$ be the eligible token set, the tokens that pass the rank
and margin test of \S\ref{sec:relaxed}; $S$ depends only on $p$, so each
$c_x$ below is fixed at a given history. We define $c_x=e^s$ for $x\in S$ and $c_x=1$ otherwise, for finite $s\geq0$.
The probability mass emitted by accepting a proposal $x$ is
\begin{equation}
 m_s(x)=\min\{q(x),c_xp(x)\},\qquad
 \alpha_s=\sum_x m_s(x).
\end{equation}
After rejection, the sampler uses the unchanged residual
\begin{equation}
 r(x)=\frac{[p(x)-q(x)]_+}{Z},
\end{equation}
where $Z=\sum_x[p(x)-q(x)]_+$ and $[z]_+=\max(z,0)$.
Thus, for $Z>0$, the next-token law is
\begin{equation}
 P_s(x)=m_s(x)+(1-\alpha_s)r(x).
 \label{eq:soft-output-law}
\end{equation}
If $Z=0$, then $p=q$, every proposal is accepted, and $P_s=p$ without
residual sampling.

\paragraph{Bound on probability amplification.}
Since $c_x\geq1$, $\alpha_s\geq\sum_x\min\{p(x),q(x)\}=1-Z$.
For $q(x)<p(x)$, $m_s(x)=q(x)$ and $1-\alpha_s\leq Z$, so
$P_s(x)\leq q(x)+p(x)-q(x)=p(x)$.
For $q(x)\geq p(x)$, the residual is zero and
$P_s(x)=m_s(x)\leq c_xp(x)$.
Together these give $P_s(x)\leq c_xp(x)\leq e^s p(x)$ for every token.
At $s=0$, Equation~\ref{eq:soft-output-law} reduces to $P_0=p$.
The bound compares distributions at the same history and for the fixed
target used in verification; it does not establish sequence-level or
task-quality equivalence to the original uncompressed model.

\paragraph{Deterministic draft proposals.}
For a draft model that deterministically proposes $y$, $q(y)=1$, so an
eligible proposal is accepted with probability $\min\{1,e^s p(y)\}$.
It is accepted unconditionally only when $p(y)\geq e^{-s}$.
This case covers the point-mass MTP proposals in the sampled verification
path; it is distinct from greedy target decoding.

\subsection{Accuracy and Acceptance Across \texorpdfstring{$s$}{s}}
\label{sec:relaxed-sweep}

\begin{table*}[t]
\centering
\small
\setlength{\tabcolsep}{4.5pt}
\begin{tabular}{@{}l rrrrr rrrr r@{}}
\toprule
 & \multicolumn{5}{c}{Accuracy (\%)} & \multicolumn{4}{c}{Accepted tokens per verify step} & \\
\cmidrule(lr){2-6} \cmidrule(lr){7-10}
Acceptance & GPQA & LCB v6 & IFBench & AIME26 & Avg.$^\dagger$ & GPQA & LCB v6 & IFBench & AIME26 & Capped$^\ddagger$ \\
\midrule
Exact & 87.4 & 83.7 & 82.0 & 99.6 & 84.4 & 2.25 & 2.15 & 2.34 & 2.30 & 7 \\
Relaxed, $s{=}0.1$ & 89.7 & 84.6 & 85.7 & 100.0 & 86.7 & 2.40 & 2.32 & 2.55 & 2.43 & 6 \\
Relaxed, $s{=}0.25$ & 88.9 & 84.3 & 84.8 & 99.2 & 86.0 & 2.60 & 2.49 & 2.78 & 2.63 & 7 \\
Relaxed, $s{=}0.5$ & 89.1 & 84.6 & 81.7 & 99.2 & 85.1 & 2.86 & 2.76 & 3.35 & 2.89 & 26 \\
Relaxed, $s{=}1$ & 87.7 & 85.7 & 76.3 & 97.5 & 83.2 & 3.25 & 3.19 & 3.72 & 3.21 & 55 \\
\bottomrule
\end{tabular}
\caption{Qwen3.8-Flash-Next with its MTP drafter (3 drafted tokens per step), three-bit weights (\S\ref{sec:weights}) and expert sharing, under each acceptance rule. Under relaxed verification, an eligible draft $y$ (in the target's top 2 and within 1.0 logit of its top token) is accepted with probability $\min(1, e^{s} p(y))$; other drafts get the exact test. 248K-token cap, thinking on, $T{=}1.0$, top-$p$ 0.95, top-$k$ 20}
\label{tab:flashnext-mtp-soft}
\end{table*}

Table~\ref{tab:flashnext-mtp-soft} sweeps $s$ on Qwen3.8-Flash-Next
with its MTP drafter, three-bit weights, and expert sharing. In short,
$s\leq0.5$ yields 6--43\% more accepted tokens per verify step with no
clear accuracy cost. The fast profile uses $s=0.5$
(Table~\ref{tab:ablation}), the most relaxed of these settings, while $s=1$ costs accuracy. The
bound in \S\ref{sec:relaxed-bound} is a worst case: it caps the factor by
which any token's probability can rise at $e^s$, about 1.11, 1.28, 1.65,
and 2.72 for the four settings, but it does not predict task accuracy,
which we measure directly.

Acceptance rises steadily with $s$. Relative to the exact rule,
accepted tokens per verify step increase by 6--9\% at $s=0.1$, 14--19\% at
$s=0.25$, 26--43\% at $s=0.5$, and 40--59\% at $s=1$, with the largest
gain on IFBench at every setting.

Accuracy holds up to $s=0.5$. At $s=0.1$, $s=0.25$, and $s=0.5$, every benchmark scores above the exact rule or within 0.4 points
of it, and the average of GPQA, LiveCodeBench, and IFBench rises from
84.4\% to 86.7\%, 86.0\%, and 85.1\%. These are means over two to eight
seeds, and the bound permits no systematic gain, so we read them as no
clear loss rather than as a gain. Accuracy first degrades at $s=1$, where IFBench falls 5.7 points and
AIME26 2.1 points, and the average drops below the exact rule. More answers
also reach the 248K-token cap as $s$ grows: 6--7 of 1,784 up to $s=0.25$,
then 26 at $s=0.5$ and 55 at $s=1$.

\section{Accuracy on Other Targets}
\label{sec:accuracy-other-targets}

Table~\ref{tab:accuracy-248k} compares the exact profile, three-bit
weights alone, and the full fast profile on Nemotron-3.5-Super,
Qwen3.8-Flash-Next, and Qwen3.8-27B.

\begin{table}[ht]
\caption{Task accuracy (\%) at a 248k-token budget. Fast is the full fast profile with relaxed verification. Avg.\ is over all five tasks, except for Flash-Next, where it leaves out the saturated AIME sets.}
\label{tab:accuracy-248k}
\centering
\small
\setlength{\tabcolsep}{3.5pt}
\begin{tabular}{@{}lrrrrrr@{}}
\toprule
 & GPQA & LCB & IFBench & AIME25 & AIME26 & Avg. \\
\midrule
\multicolumn{7}{@{}l}{\textit{Nemotron-3.5-Super}} \\
\quad Exact & 87.2 & 86.5 & 75.2 & 95.8 & 94.6 & 87.9 \\
\quad 3-bit & 86.7 & 82.6 & 71.3 & 94.2 & 95.8 & 86.1 \\
\quad Fast & 87.7 & 79.7 & 70.2 & 97.1 & 94.6 & 85.8 \\
\midrule
\multicolumn{7}{@{}l}{\textit{Qwen3.8-Flash-Next}} \\
\quad Exact & 90.7 & 87.7 & 85.3 & 100.0 & 100.0 & 87.9 \\
\quad 3-bit & 89.6 & 85.4 & 83.3 & 99.2 & 99.2 & 86.1 \\
\quad Fast & 87.0 & 86.0 & 81.7 & 97.5 & 97.5 & 84.9 \\
\midrule
\multicolumn{7}{@{}l}{\textit{Qwen3.8-27B}} \\
\quad Exact & 89.2 & 84.3 & 84.5 & 99.6 & 98.8 & 91.3 \\
\quad 3-bit & 86.0 & 87.1 & 82.3 & 89.6 & 96.2 & 88.3 \\
\quad Fast & 86.5 & 86.9 & 82.3 & 92.5 & 95.4 & 88.7 \\
\bottomrule
\end{tabular}
\end{table}

\section{Two DGX Sparks}
\label{sec:two-sparks}

Hibana can also serve one model across two DGX Sparks connected by a RoCE
link, using tensor parallelism. Attention, recurrent, and dense projection
layers are split by head or by row, and the three-bit routed experts are
split by expert, so each Spark holds half of the experts whole.
Table~\ref{tab:two-sparks} reports the resulting decode speedup on
Qwen3.8-Flash-Next, which grows from $1.56\times$ at one request to
$1.80\times$ at 32 requests.

\begin{table}[ht]
\centering
\caption{Decode throughput of Qwen3.8-Flash-Next on two DGX Sparks relative
to one. At 16 and 32 requests, Hibana uses the fast profile's optimizations
with a larger batch limit and shorter drafts.}
\label{tab:two-sparks}
\small
\begin{tabular}{rr}
\toprule
Requests & Two Sparks vs.\ one \\
\midrule
1 & $1.56\times$ \\
4 & $1.59\times$ \\
16 & $1.79\times$ \\
32 & $1.80\times$ \\
\bottomrule
\end{tabular}
\end{table}

\section{Context Reuse per Trace}
\label{sec:agent-reuse}

This appendix explains why the engines reuse different amounts of
context in the AgentPerf-Local replay (\S\ref{sec:eval-agents}). We
replay the eight default tasks on Qwen3.8-Flash-Next with one agent, so
nothing is evicted and every cache miss comes from where an engine can
resume.

\paragraph{Where an engine can resume.}
Attention layers keep one KV entry per token, so an engine can keep
any shared prefix and drop the rest. Flash-Next's recurrent layers instead
keep one running state that every token updates, so an engine can resume
only at a token where it saved a copy of that state. Each copy is about as
large as 2,992 tokens of KV cache, so engines save only a few. vLLM saves
the leading instructions, the edge of a 4,096-token prefill chunk, and the
end of each request. TensorFold resumes only from the exact end of the
previous request. In Flare's public code (commit \texttt{6fa01a49}), the pool of
saved states is fixed and shared by all requests. Each request gets an
equal share, five on Flash-Next, and when the pool fills, Flare first evicts
the saved states closest to their neighbours. Hibana also saves the start of the last message, the
point an agent's next turn usually keeps.

\begin{figure}[t]
\centering
\definecolor{exhibana}{HTML}{2978D6}\definecolor{exvllm}{HTML}{1CB07A}
\definecolor{extf}{HTML}{EB6933}\definecolor{exnote}{HTML}{D4880A}
\begin{takeaway}[Example: django-15851, turn 11 $\to$ 12]
\resizebox{\linewidth}{!}{\begin{tikzpicture}[x=1cm,y=1cm,font=\scriptsize,
  bar/.style={minimum height=0.32cm,inner sep=0pt,anchor=south west}]
\def\X#1{1.75+#1/8215*4.0}
\node[anchor=west] at (0,2.66) {Turn 11};
\node[anchor=west,text=gray] at (0,2.42) {7,789 tok};
\fill[gray!35] (1.75,2.38) rectangle ({\X{7744}},2.70);
\fill[exnote] ({\X{7744}},2.38) rectangle ({\X{7744}+0.12},2.70);
\node[anchor=west,text=exnote] at ({\X{7744}+0.15},2.54) {note};
\node at ({(1.75+\X{7744})/2},2.54) {conversation so far};
\node[anchor=west] at (0,2.06) {Turn 12};
\node[anchor=west,text=gray] at (0,1.82) {8,215 tok};
\fill[gray!35] (1.75,1.78) rectangle ({\X{7744}},2.10);
\fill[gray!15] ({\X{7744}},1.78) rectangle ({\X{8170}},2.10);
\fill[exnote] ({\X{8170}-0.06},1.78) rectangle ({\X{8215}+0.06},2.10);
\node at ({(1.75+\X{7744})/2},1.94) {same 7,744 tokens};
\node[anchor=west,text=gray] at ({\X{8215}+0.12},1.94) {+ reply, tool result};
\draw[densely dashed,gray] ({\X{7744}},0.05) -- ({\X{7744}},2.78);
\foreach \y/\name/\col/\r/\txt in {
    1.22/Hibana/exhibana/7744/re-reads 471,
    0.72/vLLM/exvllm/477/re-reads 7{,}738,
    0.22/TensorFold/extf/0/re-reads 8{,}215} {
  \node[anchor=west,text=\col] at (0,\y+0.16) {\textbf{\name}};
  \fill[\col!20] ({\X{\r}},\y) rectangle ({\X{8215}},\y+0.32);
  \ifnum\r>0 \fill[\col] (1.75,\y) rectangle ({\X{\r}},\y+0.32); \fi
  \node[anchor=west,text=\col] at ({\X{8215}+0.08},\y+0.16) {\txt};
}
\end{tikzpicture}}
\par\smallskip
\scriptsize Turn 12 drops turn 11's note and appends the reply, the tool
result, and a new note. Every end-of-request save point contains the old
note, so it cannot be used. TensorFold saves only there and re-reads the
whole prompt; vLLM falls back to the leading instructions at token 477;
Hibana also saved the start of the last message, at 7,744.
Dark: served from cache; light: recomputed.
\end{takeaway}
\caption{Where each engine resumes on one turn of a SWE-bench
trace, drawn to scale.}
\label{fig:turn-example}
\end{figure}
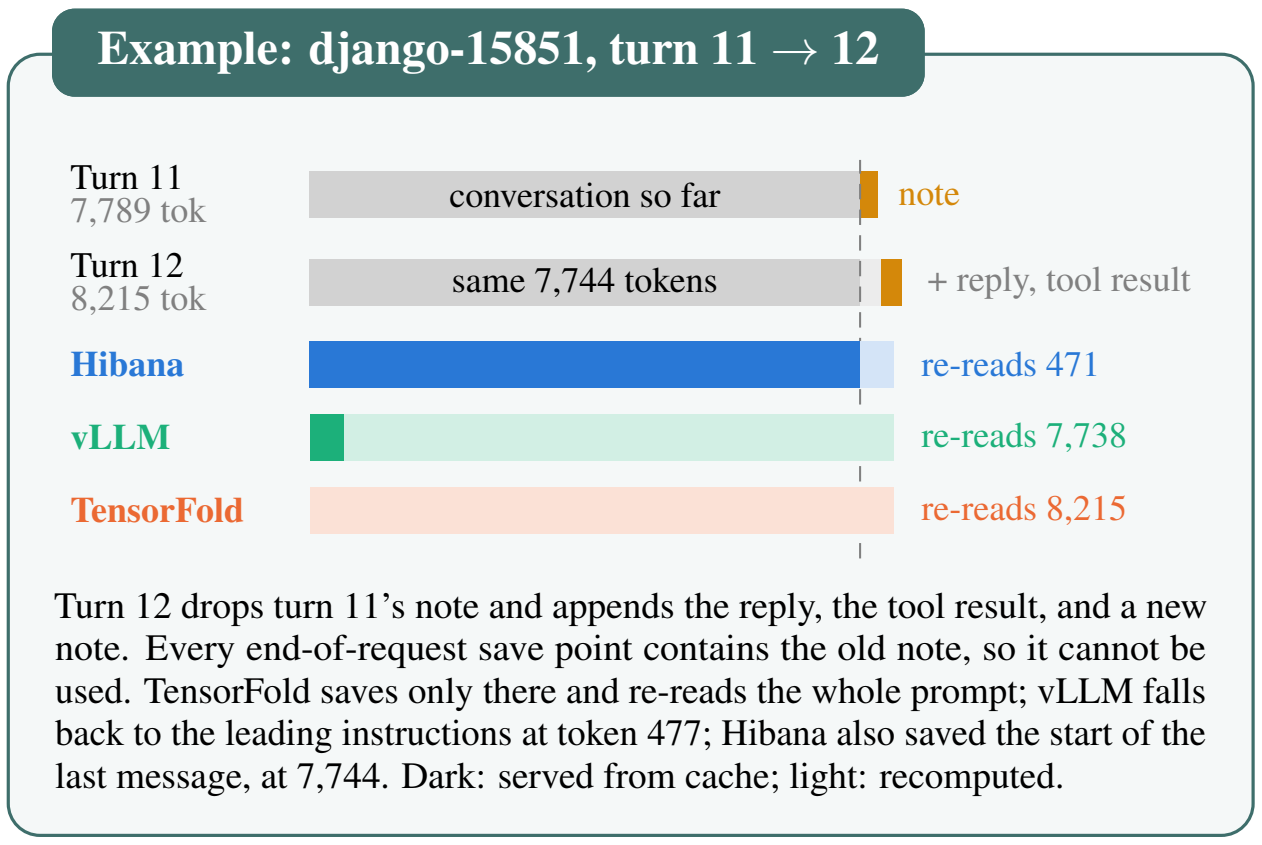

\begin{figure*}[t]
\centering
\includegraphics[width=\linewidth]{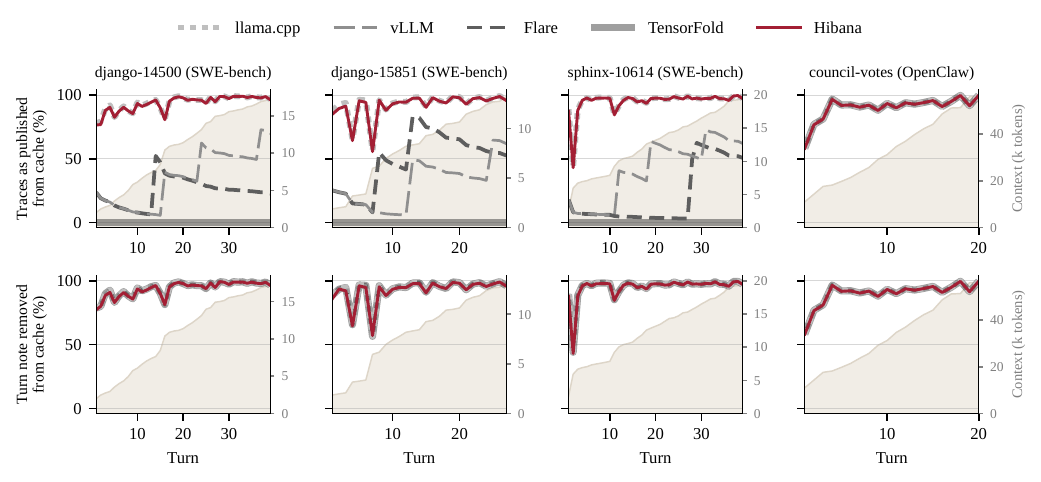}
\caption{Share of each prompt served from cache on every turn of
four AgentPerf-Local tasks, Qwen3.8-Flash-Next, one agent. Gray: the
prompt's length. Top: traces as published, where each SWE-bench request
ends with a turn note that the next turn deletes. Bottom: the same traces
with the note removed. The first turn of each task is omitted.
Flare runs its recipe defaults; Table~\ref{tab:agent-reuse}
omits llama.cpp.}
\label{fig:agent-reuse}
\end{figure*}

\paragraph{The SWE-bench turn note.}
The three SWE-bench tasks end every request with a ``Tool-call turn
$k$ of 100'' message and delete it on the next turn; the note is part of
the published traces. Every end-of-request save point therefore includes a
note that the next prompt no longer has.
Figure~\ref{fig:turn-example} walks through one such turn, and
Figure~\ref{fig:agent-reuse} (top) shows the result on every turn.
TensorFold serves nothing from cache, vLLM's hit rate decays as the
context grows and jumps only when its chunk-edge point moves up by 4,096
tokens, and Hibana stays at 85--98\%. With only five
checkpoints per request, Flare, like vLLM, often has none near the
prompt end, so it falls back thousands of tokens and jumps only when one
lands close. With room in its pool it does keep the boundary before the
note: a fresh Nemotron-3.5-Bolt server serves 94--99\% of django-14500
from cache, and 86\% once the eight tasks fill the pool and evict that
checkpoint. With 32 agents sharing the pool (the scaling run of Figure~\ref{fig:speed-eval}b, a different run from the one-agent table below), Flare serves 41\% of these
prompts from cache on Nemotron-3.5-Bolt and 19\% on Qwen3.8-Flash-Next. At 32 agents its GPU is fully busy but
decodes only 19--29\% of the time, and it recomputes 1.5--2.1$\times$
vLLM's prompt tokens (Qwen3.8-27B, Nemotron-3.5-Bolt, and Nemotron-3.5-Super).
In our runs, llama.cpp stays close to Hibana. On the
append-only OpenClaw tasks, the engines reuse similar amounts of context on the three longer tasks. On the two five-turn tasks, TensorFold caches more (92\% against 78\% on \texttt{k8s-debugging}), and on \texttt{skill-search} Hibana and TensorFold both do (94\% and 96\% against 78\%) (Table~\ref{tab:agent-reuse}).

\begin{table*}[t]
\centering
\caption{Context reuse per task, traces as published,
Qwen3.8-Flash-Next, one agent. Re-read is the prompt tokens an engine
recomputes per turn, excluding each task's first turn.}
\label{tab:agent-reuse}
\small
\setlength{\tabcolsep}{4.5pt}
\begin{tabular}{llrrrrrrrrr}
\toprule
 & & & \multicolumn{4}{c}{Served from cache (\%)} & \multicolumn{4}{c}{Re-read per turn (tokens)} \\
\cmidrule(lr){4-7} \cmidrule(lr){8-11}
Task & Kind & Turns & vLLM & Hibana & TensorFold & Flare & vLLM & Hibana & TensorFold & Flare \\
\midrule
django-14500 & SWE-bench & 40 & 44 & 95 & 0 & 26 & 6,018 & 473 & 10,718 & 7,893 \\
django-15851 & SWE-bench & 28 & 36 & 93 & 0 & 56 & 5,267 & 499 & 8,210 & 3,595 \\
sphinx-10614 & SWE-bench & 40 & 49 & 96 & 0 & 26 & 6,482 & 507 & 12,708 & 9,345 \\
council-votes & OpenClaw & 21 & 93 & 92 & 93 & 93 & 2,483 & 2,520 & 2,490 & 2,483 \\
council-budget & OpenClaw & 20 & 91 & 91 & 91 & 91 & 1,991 & 2,020 & 1,998 & 1,991 \\
searchable-index & OpenClaw & 9 & 87 & 87 & 87 & 87 & 2,784 & 2,825 & 2,791 & 2,784 \\
k8s-debugging & OpenClaw & 5 & 78 & 78 & 92 & 78 & 410 & 446 & 417 & 410 \\
skill-search & OpenClaw & 5 & 78 & 94 & 96 & 78 & 281 & 316 & 288 & 281 \\
\bottomrule
\end{tabular}
\end{table*}

\paragraph{Removing the per-turn prompt edit}
To confirm that the note alone causes these misses, we replayed the
same traces with the 108 note messages deleted, so every task only
appends. All engines then follow the same line
(Figure~\ref{fig:agent-reuse}, bottom): vLLM caches 93\% of each prompt
over all eight tasks and TensorFold 86--93\%, while Hibana stays at 92\%.
Flare rises from 26--56\% to 93--95\% on the SWE-bench tasks.
Hibana still completes about $2\times$ as many tasks per hour as vLLM (77
against 37 at one agent, 154 against 82 at 16), and this remaining gap
comes from decoding speed: about 113 tokens/s against vLLM's 46 at one
agent, from a fit of each turn's time to its output tokens and uncached prompt
tokens. The same fit puts prefill at 13\% of vLLM's one-agent time and 16\% of
Hibana's, so decoding explains almost all of the gap. Flare's tasks per hour at 32 agents rise 1.7--2.5$\times$
(37 to 80 on Flash-Next).
Without the note, vLLM's cache share on Flash-Next rises from 70\% to 93\% and its tasks per hour at 16 agents from 49 to 82, while the other three targets are unchanged. Setup details. The replay runs AgentPerf-Local's eight recorded tasks (168 turns) in a closed loop. Each of $N$ agents takes one task and sends its turns back to back with no tool time. Every request carries the recorded conversation and tool schemas, with thinking off, and generates the recorded output length, so all engines do the same work on token-identical prompts. We discard each engine's reply, so a task here is a replayed task, not a solved one. Tasks per hour is tasks finished over wall time, and at eight or more agents the slowest task sets the wall time. On Flash-Next, vLLM and Flare serve NVIDIA's FP4-Local checkpoint and Hibana serves RadixArk's NVFP4 build. The replay uses Hibana's throughput profile, which keeps exact verification only on Nemotron-3.5-Bolt. At 32 agents on Flash-Next, vLLM's cache share falls to 29\% and its 7.4M uncached prompt tokens take most of a 4,027\,s run, against 0.8M for Hibana. The replay differs from Artificial Analysis's own setting, which leaves thinking at the server default. The burst study's TTFT counts only replies streamed in more than three chunks over at least 50\,ms, because Hibana returns tool-call replies in one piece.